\documentclass[10pt,letterpaper]{article}
\usepackage[top=1in,bottom=1in,left=1in,right=1in]{geometry}

\usepackage[utf8]{inputenc}
\usepackage[T1]{fontenc}
\usepackage{lmodern}

\usepackage{cite}

\usepackage{nameref,hyperref}
\hypersetup{
    citebordercolor={0.5 0.25 0.7}, 
    linkbordercolor={1 1 1}, 
    urlbordercolor={0.5 0.25 0.7}  
}

\usepackage[right]{lineno}

\usepackage{microtype}
\DisableLigatures[f]{encoding = *, family = * }

\usepackage{changepage}

\usepackage[aboveskip=1pt,labelfont=bf,labelsep=period,singlelinecheck=off,textfont=it]{caption}

\makeatletter
\renewcommand{\@biblabel}[1]{\quad#1.}
\makeatother

\usepackage{lastpage,fancyhdr,graphicx}
\usepackage{epstopdf}
\usepackage{color}

\definecolor{Gray}{gray}{.25}

\usepackage{graphicx}

\usepackage{sidecap}

\usepackage{wrapfig}
\usepackage[pscoord]{eso-pic}
\usepackage[fulladjust]{marginnote}
\reversemarginpar

\usepackage{amsmath}
\usepackage{amssymb}
\usepackage{float}

\graphicspath{{images/}}

\begin{document}

\begin{flushleft}
{\Large
\textbf{Landslide Detection and Mapping Using Deep Learning Across Multi-Source Satellite Data and Geographic Regions}
}
\\[0.15in]
Dr. Rahul A. Burange\textsuperscript{*},
Harsh K. Shinde\textsuperscript{*},
Omkar Mutyalwar
\\
\bigskip
Department of Electronics \& Telecommunication, KDK College of Engineering, Nagpur, India

\end{flushleft}

\section*{Abstract}
\addcontentsline{toc}{section}{Abstract}
Landslides pose severe threats to infrastructure, economies, and human lives, necessitating accurate detection and predictive mapping across diverse geographic regions. With advancements in deep learning and remote sensing, automated landslide detection has become increasingly effective. This study presents a comprehensive approach integrating multi-source satellite imagery and deep learning models to enhance landslide identification and prediction. We leverage Sentinel-2 multispectral data and ALOS PALSAR-derived slope and Digital Elevation Model (DEM) layers to capture critical environmental features influencing landslide occurrences. Various geospatial analysis techniques are employed to assess the impact of terrain characteristics, vegetation cover, and rainfall on detection accuracy. Additionally, we evaluate the performance of multiple state-of-the-art deep learning segmentation models, including U-Net, DeepLabV3+, and Res-Net, to determine their effectiveness in landslide detection. The proposed framework contributes to the development of reliable early warning systems, improved disaster risk management, and sustainable land-use planning. Our findings provide valuable insights into the potential of deep learning and multi-source remote sensing in creating robust, scalable, and transferable landslide prediction models.

\bigskip
\textbf{\textit{Index Terms:}} \textit{Image Processing, Machine Learning, Deep Learning, Computer Vision, Remote Sensing.}


\section{Introduction}
Landslides represent a significant natural hazard, causing substantial environmental and socio-economic damage worldwide. The increasing frequency of extreme weather events, deforestation, and rapid urbanization have exacerbated the risks associated with landslides, highlighting the need for effective detection and monitoring strategies \cite{guzzetti2006, guzzetti2022}. Traditional landslide mapping techniques, including field surveys and manual interpretation of satellite imagery, are time-consuming, costly, and often constrained by limited spatial coverage \cite{cruden1996, galli2008}. Consequently, the integration of remote sensing technologies and machine learning-based approaches has gained prominence in recent years to automate and enhance landslide detection and segmentation \cite{lima2017, ghorbanzadeh2019}.

Remote sensing plays a critical role in landslide detection, offering diverse data sources, such as optical imagery, synthetic aperture radar (SAR), and digital elevation models (DEMs) \cite{noviello2020, pradhan2015}. Optical satellite sensors, such as Sentinel-2 and Landsat-8, provide valuable spectral information that helps identify vegetation changes and surface disturbances associated with landslides. However, these sensors are susceptible to cloud cover and seasonal variations, which can obstruct land surface visibility. SAR data, acquired from platforms such as Sentinel-1 and ALOS PALSAR, overcome these limitations by offering all-weather and day-night imaging capabilities, making them particularly valuable for monitoring terrain deformation and slope instability. The fusion of optical and SAR data has been shown to improve landslide detection accuracy by capturing both spectral and structural characteristics of affected areas. Additionally, high-resolution aerial and UAV imagery has been increasingly utilized to provide detailed, real-time insights into landslide-prone regions \cite{dai2020}.

Advancements in deep learning and computer vision have significantly improved landslide segmentation performance \cite{azmoon2021, kaushal2024}. Traditional machine learning techniques, such as support vector machines (SVMs) and random forests (RFs), rely on handcrafted features, which require expert knowledge and are less adaptable to varying geographic conditions. Deep learning models, particularly convolutional neural networks (CNNs) and transformer-based architectures, have demonstrated superior capabilities in feature extraction and spatial dependency learning from complex datasets. Models such as U-Net, DeepLabV3+, and SegFormer have gained widespread adoption for landslide segmentation due to their ability to capture fine-grained spatial patterns while maintaining computational efficiency \cite{kaushal2024}. Furthermore, hybrid architectures, such as HRNet and attention-based networks, have further enhanced segmentation accuracy by incorporating global contextual information.

Despite these advancements, several challenges persist in real-world landslide detection and segmentation. One primary issue is the imbalanced distribution of landslide data, as landslide occurrences are relatively rare compared to stable terrain. This imbalance can lead to biased model predictions, reducing overall classification accuracy \cite{titti2021}. Additionally, variations in landslide morphology due to differences in soil composition, slope gradient, and vegetation cover make it difficult to generalize models across diverse geographic regions. Another major challenge arises from noise and occlusions in satellite imagery, such as cloud interference, seasonal vegetation changes, and sensor limitations, which introduce uncertainties in model predictions. To mitigate these challenges, researchers have explored various strategies, including data augmentation, domain adaptation, and ensemble learning, to improve model robustness and generalization.

The selection of appropriate loss functions and training strategies is crucial for optimizing deep learning models for landslide detection. Commonly used loss functions, such as cross-entropy loss and Dice loss, may struggle with class imbalance. Alternative loss functions, including focal loss and IoU-based loss, have been introduced to emphasize hard-to-classify regions and enhance segmentation accuracy. Additionally, training optimization techniques, such as learning rate scheduling, transfer learning, and self-supervised learning, have been employed to accelerate model convergence and improve generalization. The integration of multi-modal data fusion, combining optical, SAR, and topographic information, has further enhanced landslide detection performance by leveraging complementary data sources.

Looking ahead, the future of landslide segmentation research is expected to focus on improving model generalization, enhancing real-time processing capabilities, and integrating multi-source geospatial data. The increasing availability of high-resolution satellite imagery, coupled with advancements in edge computing and cloud-based geospatial platforms, will enable more scalable and efficient landslide monitoring systems \cite{khaing2020}. Furthermore, the integration of physics-informed machine learning models, which incorporate domain knowledge from geotechnical and hydrological studies, holds promise for improving model interpretability and reliability \cite{yang2022}. The development of open-source datasets and standardized benchmarking frameworks will be instrumental in advancing the field, facilitating fair comparisons, and accelerating innovation in landslide segmentation methodologies \cite{ghorbanzadeh2022}.

In summary, the integration of remote sensing and deep learning has led to significant progress in landslide detection and segmentation. While challenges such as data imbalance, model generalization, and noise in satellite imagery persist, ongoing research efforts continue to refine methodologies and improve accuracy. The combination of multi-source data, advanced deep learning architectures, and robust training strategies will be critical in developing more reliable and scalable landslide detection systems. As research in this domain evolves, deep learning-driven landslide segmentation is expected to play a crucial role in disaster risk management, early warning systems, and environmental monitoring, contributing to more effective and data-driven decision-making in landslide-prone regions.

\section{Dataset Description}
This study aims to devise a multi-source landslide benchmark dataset that addresses the need for annotated data to train deep learning (DL) models for landslide detection. The amount of data required to achieve optimal performance in supervised DL models is not clearly defined, but it has been established that small training datasets of labeled images lead to poor classification performance. Even large training datasets may not encompass all possible scenarios, which limits model generalization. This issue is particularly critical in landslide detection, as landslides vary in size, shape, and geographical location. To mitigate this, we selected landslide-affected study areas from four different geographical regions to enhance dataset diversity. The landslides in these areas were triggered by different factors, sometimes in combination, which can improve the transferability of trained models.

Deep Learning models require a large quantity of labeled data to efficiently learn various parameters. According to research, when only a small training dataset of labeled images is used, classification performance may degrade, while a large training dataset cannot cover all conceivable cases. To resolve this problem, we used a benchmark dataset called Landslide4Sense, which consists of study sites affected by landslides from diverse geographical areas \cite{ghorbanzadeh2022}.

It provides information about the elevation of the terrain, while details on the steepness of the terrain are offered by the slope data, both of which are important factors in the analysis of landslide susceptibility. The Landslide4Sense benchmark collection contains 14 layers of data, including Sentinel-2 multispectral data (bands 1-12) and ALOS PALSAR slope data. This benchmark dataset’s 14 layers have been thoroughly labeled for landslide and non-landslide classifications, with each layer resized to approximately 10m per pixel resolution. This dataset includes occurrences from multiple regions and is divided into three subsets: train, validation, and test sets. This allows for enhanced training of DL models, helping them handle novel circumstances or out-of-distribution inputs more successfully and improving their performance across regions. The training group is constructed from four locations: Hokkaido’s Iburi-Tobu, Karnataka’s Kodagu, Bagmati’s Rasuwa, and western Taitung.

The data obtained was used to create 3799 patches, each measuring $128\times128$ pixels. Furthermore, the validation and test sets contain 245 and 800 equal-sized image patches, respectively. Ranging in wavelength from ultra blue to short-wave infrared (SWIR), Sentinel-2 provides multispectral layers. Bands B2, B3, B4, and B8 have a spatial resolution of 10 meters, whereas bands B5, B6, B7, B11, and B12, as well as B1, B9, and B10, have resolutions of 20 meters and 60 meters, respectively. The Alaska Satellite Facility offers a high-resolution DEM from the ALOS PALSAR, and the slope layer is produced from this DEM. Both the DEM and the slope layers are transformed to a spatial resolution of 10 meters.

\begin{figure}[H]
\centering
\includegraphics[width=\textwidth]{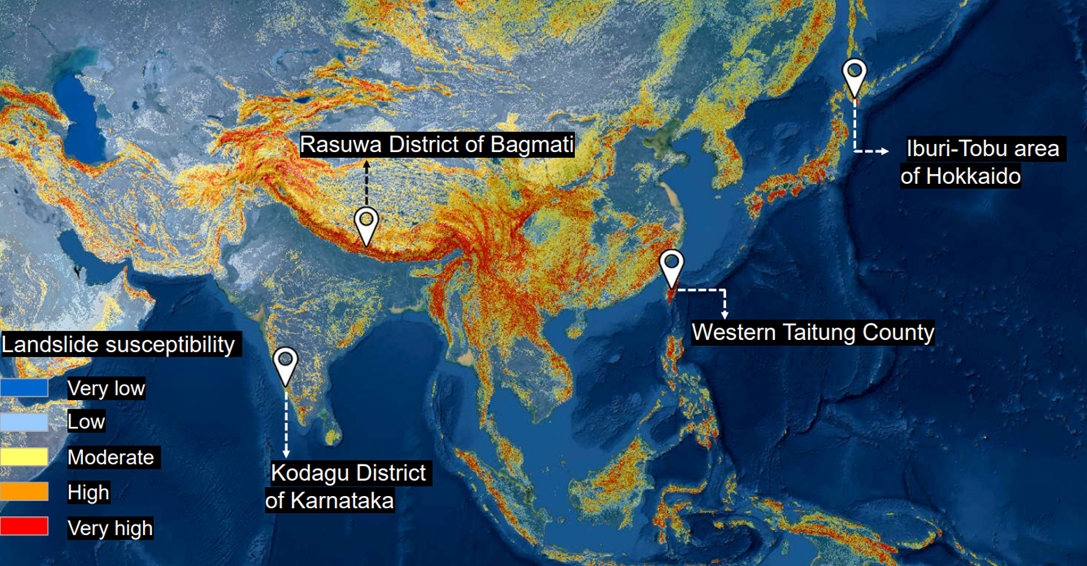}
\caption{\color{Gray} \textbf{Locations of the training sites on a global image retrieved from \cite{ghorbanzadeh2022} for landslide susceptibility.}}
\label{fig1}
\end{figure}

\subsection{Study Areas}
The selected study areas represent diverse geographic and climatic conditions. Their locations are depicted on a global landslide susceptibility map generated using multiple explanatory variables such as slope degree, forest loss, geology, road networks, and fault lines.

\paragraph{Iburi-Tobu Area of Hokkaido, Japan:}
Hit by a magnitude 6.6 earthquake on September 6, 2018, triggering over 5600 landslides. Landslides were exacerbated by preceding heavy rainfall from Typhoon Jebi. Landslide inventories were created using very high-resolution aerial images.
\begin{figure}[H]
\centering
\includegraphics[width=0.48\textwidth]{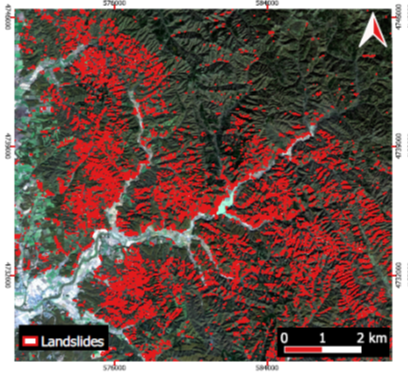} \hfill
\includegraphics[width=0.48\textwidth]{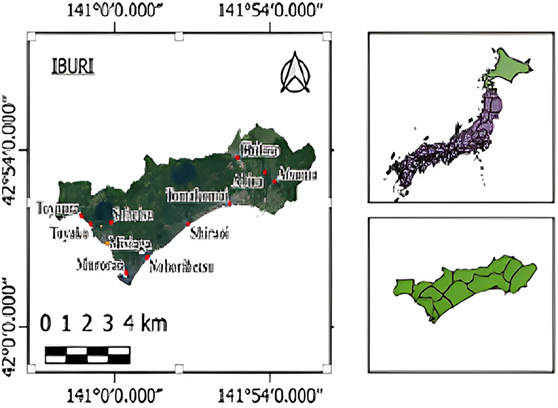}
\caption{\color{Gray} \textbf{Iburi-Tobu} from \cite{ghorbanzadeh2022, kaushal2024}.}
\label{fig2}
\end{figure}

\paragraph{Kodagu District of Karnataka, India:}
Experienced extreme rainfall in August 2018, triggering severe landslides and flash floods. Landslides were linked to deforestation, unplanned urbanization, and mining activities. Previous studies have applied unsupervised learning techniques for landslide detection.
\begin{figure}[H]
\centering
\includegraphics[width=0.48\textwidth]{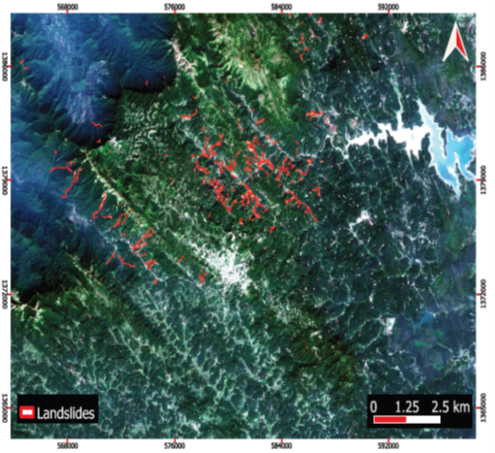} \hfill
\includegraphics[width=0.48\textwidth]{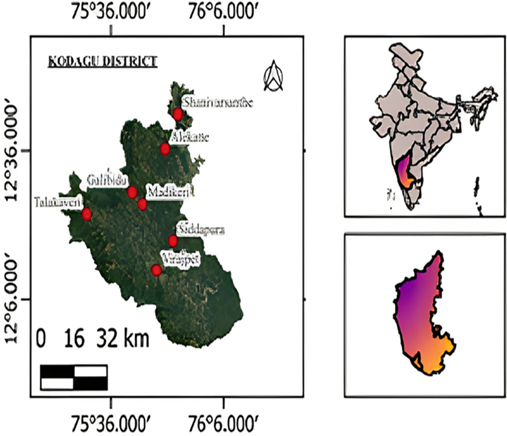}
\caption{\color{Gray} \textbf{Kodagu} from \cite{ghorbanzadeh2022, kaushal2024}.}
\label{fig3}
\end{figure}

\paragraph{Rasuwa District of Bagmati, Nepal:}
One of the most landslide-prone regions in the Himalayas. Major landslides occurred due to the 2015 Gorkha and Dolakha earthquakes. Landslide inventory compiled from GPS field surveys and visual interpretation of high-resolution images.
\begin{figure}[H]
\centering
\includegraphics[width=0.4\textwidth]{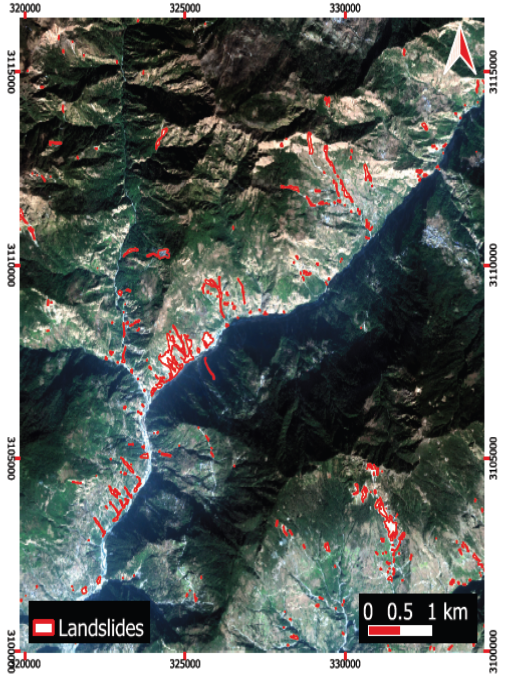} \hfill
\includegraphics[width=0.4\textwidth]{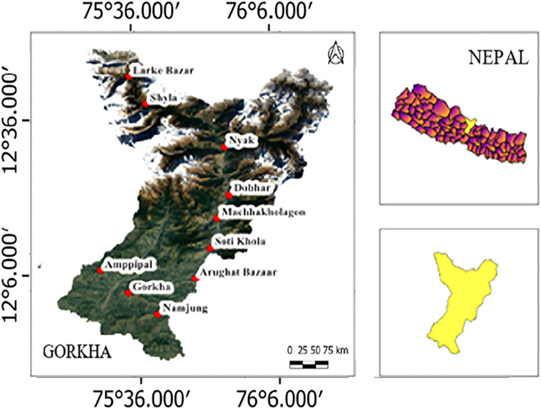}
\caption{\color{Gray} \textbf{Rasuwa \& Gorkha District - Nepal} from \cite{ghorbanzadeh2022, kaushal2024}.}
\label{fig4}
\end{figure}

\paragraph{Western Taitung County, Taiwan:}
Landslides frequently triggered by typhoons and earthquakes. Typhoon Morakot (2009) caused extensive landslides, destroying villages and infrastructure. Landslide inventory derived from previous studies and Google Earth images.
\begin{figure}[H]
\centering
\includegraphics[width=0.4\textwidth]{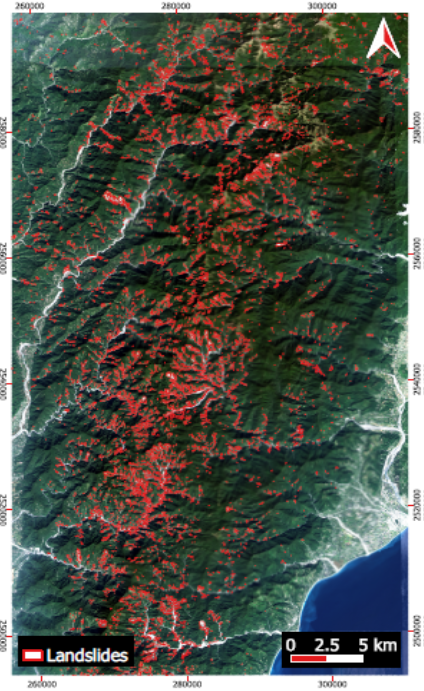} \hfill
\includegraphics[width=0.4\textwidth]{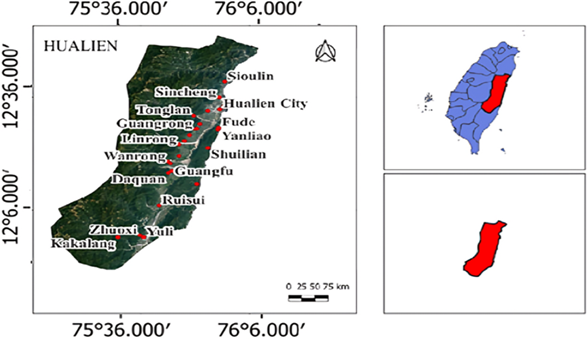}
\caption{\color{Gray} \textbf{Western Taitung County \& Hualien – Taiwan} from \cite{ghorbanzadeh2022, kaushal2024}.}
\label{fig5}
\end{figure}

\subsection{Sensor Characteristics}
\textbf{Sentinel-2:} Provides multi-spectral imagery with 13 bands at spatial resolutions of 10, 20, and 60 meters. High revisit frequency (2–3 days at mid-latitudes) enables continuous monitoring. Sentinel-2 data were obtained from Google Earth Engine (GEE), ensuring cloud-free imagery for analysis.

\textbf{ALOS PALSAR:} Provides synthetic aperture radar (SAR) data with a 12.5m spatial resolution. The DEM and slope layers derived from ALOS PALSAR were used to supplement optical imagery.

\subsection{Landslide Inventory Annotation}
To ensure high-quality landslide annotations, we employed a two-step workflow:
\begin{itemize}
    \item \textbf{Object-Based Image Analysis (OBIA):} Computed image difference indices from pre- and post-landslide images. Performed multi-resolution segmentation and rule-based classification.
    \item \textbf{Manual Verification:} Used high-resolution Google Earth imagery and existing landslide inventories. Ensured accurate delineation of landslide boundaries.
\end{itemize}

\subsection{Benchmark Dataset Statistics and Structure}
The Landslide4Sense benchmark dataset includes $128\times128$ window-size patches, each containing 14 distinct data layers. The first 12 bands consist of multi-spectral data from Sentinel-2, while bands 13 and 14 represent the Digital Elevation Model (DEM) and slope data derived from ALOS PALSAR. Each patch is accurately labeled, with ground truth polygons outlined in red to indicate landslide areas. These labeled patches provide essential annotations for training and evaluating deep learning models, ensuring precise classification of landslide-prone regions.

\begin{figure}[H]
\centering
\includegraphics[width=0.85\textwidth]{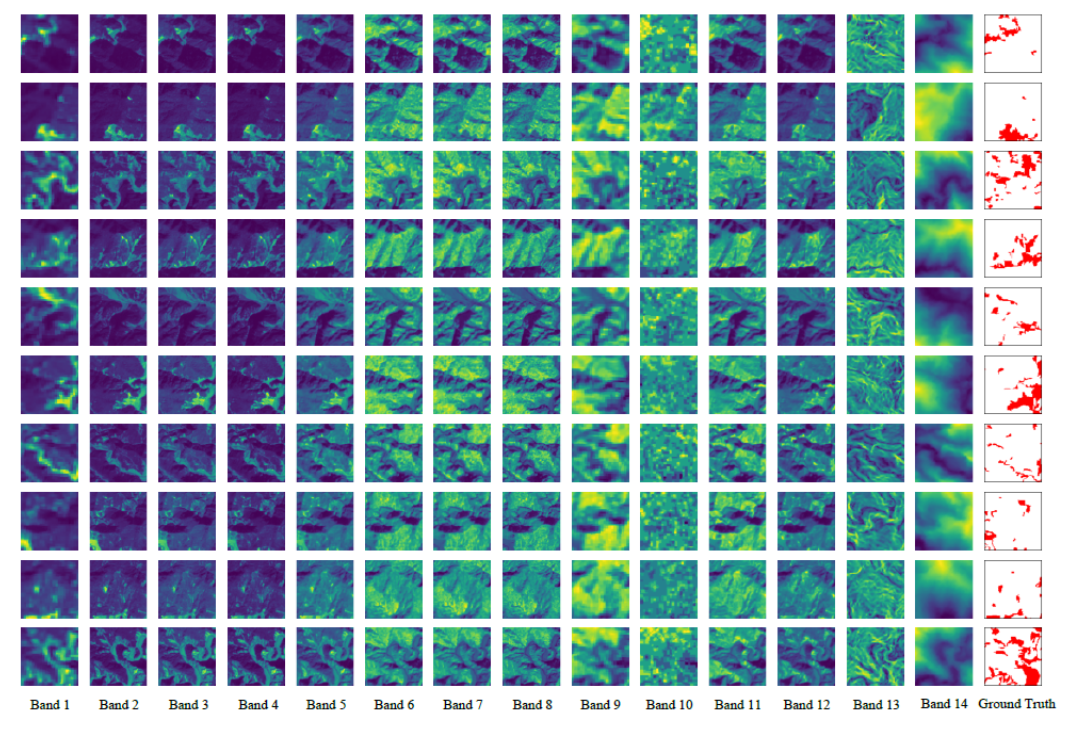}
\caption{\color{Gray} \textbf{Visualizing each unique layer inside the generated landslide dataset’s 128x128 window-size patches.} The first 12 bands feature multi-spectral data from Sentinel-2, while bands 13 and 14 contain DEM data and slope from ALOS PALSAR. The patches in the last column are accurately labeled, and they are complimented by red polygons signifying the landslide category. These patches in the last column refer to the ground truth Polygons. Dataset image copyright © Omid Ghorbanzadeh. \cite{ghorbanzadeh2022}}
\label{fig6}
\end{figure}

The 14 layers in the Landslide4Sense dataset are described below:
\begin{itemize}
    \item Sentinel-2 band 1: Blue spectral band data.
    \item Sentinel-2 band 2: Green spectral band data.
    \item Sentinel-2 band 3: Red spectral band data.
    \item Sentinel-2 band 4: Near Infrared (NIR) spectral band data.
    \item Sentinel-2 band 5: Shortwave Infrared (SWIR) spectral band data.
    \item Sentinel-2 band 6: Shortwave Infrared (SWIR) spectral band data.
    \item Sentinel-2 band 7: Shortwave Infrared (SWIR) spectral band data.
    \item Sentinel-2 band 8: NIR spectral band data.
    \item Sentinel-2 band 9: Water Vapour (WV) spectral band data.
    \item Sentinel-2 band 10: Cirrus (CI) spectral band data.
    \item Sentinel-2 band 11: SWIR spectral band data.
    \item Sentinel-2 band 12: SWIR spectral band data.
    \item Digital Elevation Model (DEM): Elevation information data.
    \item Slope: Slope information data.
\end{itemize}

The dataset includes 3799 annotated image patches ($128 \times 128$ pixels each, with a resolution of 10 meters per pixel), and a dataset split of 959 patches for training and 2840 patches for testing (or training-3799, validation-245, testing-800 depending on the split logic). Each patch contains pixel-wise labels indicating landslide and non-landslide areas. The dataset exhibits significant variability in landslide shape, size, distribution, and frequency across study areas. Sentinel-2 bands 4 and 5 show the highest spectral differences between landslide and non-landslide areas. Elevation and slope data provide additional distinguishing features, particularly in regions with steep terrain. This dataset serves as a valuable resource for training and evaluating DL models in landslide detection, providing diverse and challenging real-world conditions to enhance model generalization and transferability.

\section{Methodology}
In this research, we focus on developing a hybrid deep learning (DL) methodology for improving landslide prediction accuracy. Landslides pose significant risks to human lives and infrastructure, necessitating precise predictive models. Traditional approaches often struggle to extract detailed spatial information from satellite imagery, motivating our adoption of advanced deep learning techniques. Our methodology leverages multiple state-of-the-art models, each contributing unique strengths to landslide detection and segmentation.

We employ U-Net, a widely used segmentation model known for its effectiveness in pixel-wise classification, alongside DeepLabV3+, a powerful semantic segmentation model. These models are paired with encoder backbones such as DenseNet121, EfficientNet-B0, InceptionResNetV2, InceptionV4, MiT-B1, MobileNetV2, ResNet34, ResNeXt50\_32X4D, SeResNet50, SeResNeXt50\_32X4D, SegFormerB2, and VGG-16. The models are trained on the Landslide4Sense dataset, which integrates Sentinel-2 multispectral data with ALOS PALSAR-derived elevation and slope information.

Our approach includes a pyramid pooling layer to capture multi-scale spatial features, enhancing the models' ability to detect landslides across different resolutions. Additionally, Object-Based Image Analysis (OBIA) is incorporated to analyze coherent objects based on spectral, textural, and contextual attributes rather than individual pixels, improving the identification of landslide-prone areas.

To ensure consistency and optimize performance, we apply data preprocessing techniques such as normalization and noise reduction. Furthermore, data augmentation is employed during training to increase dataset diversity and improve model generalization.

This research advances landslide prediction capabilities by integrating multiple deep learning architectures, providing valuable insights for proactive disaster management and mitigation strategies in landslide-prone regions.

\begin{table}[!ht]
\centering
\caption{{\bf Description of the Landslide4Sense Dataset.}}
\begin{tabular}{|c|l|p{8cm}|}
\hline
\textbf{Serial No.} & \textbf{Attribute} & \textbf{Description} \\ \hline
1 & Dataset Name & Landslide4Sense \\ \hline
2 & Number of Samples & Training - 3799, Testing - 800, Validation - 245 \\ \hline
3 & Target Variable & Landslide (0: No landslide, 1: Landslide) \\ \hline
4 & Data Source & Landslide monitoring multi-sensors \\ \hline
5 & Geographic Coverage & Bagmati’s Rasuwa district, Karnataka’s Kodagu district, Hokkaido’s Iburi-Tobu area, Taiwan’s western Taitung County \\ \hline
6 & Preprocessing & Removal of missing values, normalization \\ \hline
7 & Feature Types & Meteorological, Geotechnical, Geological, Topographic \\ \hline
8 & Feature Granularity & Temporal and spatial averages or measurements \\ \hline
9 & Data Format & Hierarchical Data Format version 5 (HDF5) \\ \hline
\end{tabular}
\label{tab1}
\end{table}

\subsection{U-Net}
U-Net is a fully convolutional neural network (FCN) designed for image segmentation tasks, making it well-suited for landslide detection. It follows an encoder-decoder architecture, where the encoder captures spatial context, and the decoder restores spatial resolution for precise segmentation. U-Net is particularly effective for remote sensing applications due to its ability to learn multi-scale features and retain fine-grained spatial information through skip connections.

In our implementation, U-Net is trained on multi-source satellite imagery with six input channels: RGB bands, Normalized Difference Vegetation Index (NDVI), slope, and elevation. The network consists of a contracting path (downsampling) with convolutional and max-pooling layers, and an expansive path (upsampling) with transposed convolutions and concatenation layers. The final segmentation mask is obtained using a $1\times1$ convolution with a sigmoid activation function.

For optimization, we use the Adam optimizer and binary cross-entropy loss, along with Dice coefficient-based metrics to evaluate model performance. The model is trained with early stopping and checkpointing to ensure optimal convergence.

\begin{figure}[H]
\centering
\includegraphics[width=\textwidth]{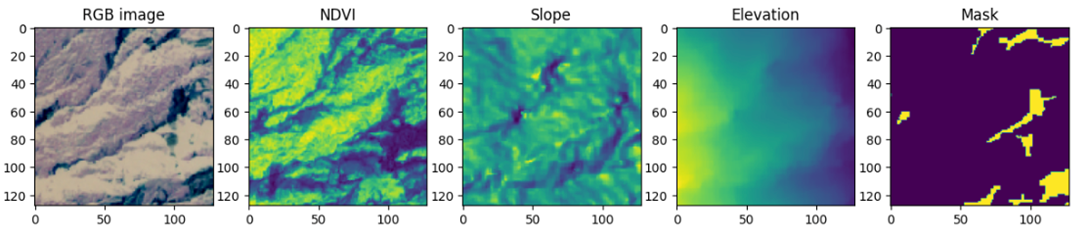}
\caption{\color{Gray} \textbf{Visual representation of U-Net training dataset images.}}
\label{fig7}
\end{figure}

\begin{figure}[H]
\centering
\includegraphics[width=\textwidth]{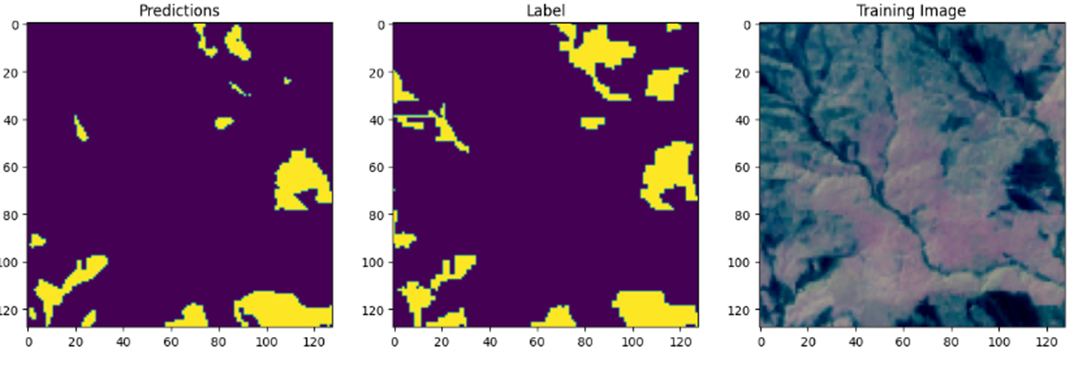}
\caption{\color{Gray} \textbf{Visual representation of U-Net Validation dataset images.}}
\label{fig8}
\end{figure}

\textbf{Mathematical Formulation of U-Net:}
\begin{enumerate}
    \item \textbf{Encoder (Contracting Path)}: Each encoder block consists of two convolutional layers with ReLU activation followed by max pooling:
    \begin{equation}
    X_{l+1} = \text{MaxPool}(f(W_l \ast X_l + b_l))
    \end{equation}
    where $X_l$ is the feature map at layer $l$, $W_l$ and $b_l$ are the weights and bias, $f(\cdot)$ is the ReLU activation function, $\ast$ denotes convolution, and $\text{MaxPool}(\cdot)$ is the max pooling operation.
    
    \item \textbf{Bottleneck}: At the bottleneck, the deepest part of the network, convolutional layers further refine features:
    \begin{equation}
    X_b = f(W_b \ast X_{b-1} + b_b)
    \end{equation}
    
    \item \textbf{Decoder (Expansive Path)}: The decoder upsamples the feature maps using transposed convolutions and concatenates with corresponding encoder feature maps:
    \begin{equation}
    X_{d+1} = f(W_d \ast \text{UpSample}(X_d) + b_d)
    \end{equation}
    Skip connections concatenate feature maps from the encoder:
    \begin{equation}
    X_{\text{skip}} = \text{Concat}(X_d, X_e)
    \end{equation}
    where $\text{UpSample}(X_d)$ is transposed convolution, $X_d$ is the decoder feature map, and $X_e$ is the corresponding encoder feature map.
    
    \item \textbf{Output Layer (Segmentation Map)}: The final segmentation map is obtained using a $1\times1$ convolution followed by a sigmoid activation function:
    \begin{equation}
    \hat{Y} = \sigma(W_o \ast X_o + b_o)
    \end{equation}
    where $\hat{Y}$ is the predicted binary mask, and $\sigma(\cdot)$ is the sigmoid activation function.
\end{enumerate}

\newpage
\subsection{DeepLabV3+}
DeepLabV3+ is an advanced deep learning model for semantic segmentation, designed to capture detailed spatial information using an encoder-decoder architecture. The encoder, based on a ResNet50 backbone, extracts multi-scale features, while the Atrous Spatial Pyramid Pooling (ASPP) module enhances contextual understanding by applying multiple parallel dilated convolutions. The decoder then refines these extracted features through upsampling and convolution layers to produce a high-resolution segmentation map. In this implementation, the model is trained for landslide detection using a combination of Weighted Binary Cross-Entropy (WCE) and Dice loss, ensuring accurate pixel-wise classification.

\begin{figure}[H]
\centering
\includegraphics[width=0.7\textwidth]{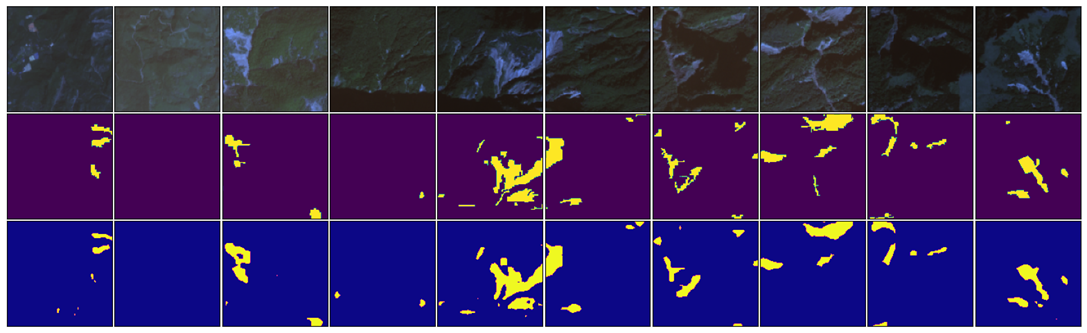}
\caption{\color{Gray} \textbf{Visual representation of DeepLabV3+}, where RGB, GroundTruth Mask (viridis) \& Prediction (plasma).}
\label{fig9}
\end{figure}

\subsection{DenseNet121}
DenseNet121 is a deep convolutional neural network known for its dense connectivity pattern, where each layer is directly connected to all its preceding layers. This architecture enhances gradient flow and reduces the number of parameters compared to traditional deep networks.
\begin{figure}[H]
\centering
\includegraphics[width=0.7\textwidth]{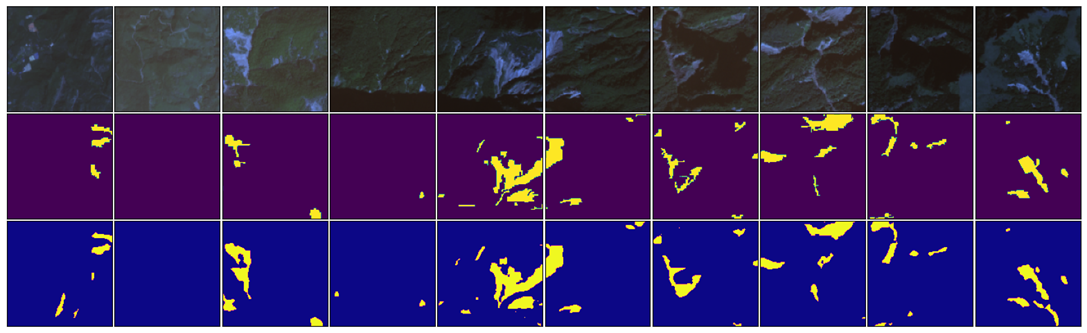}
\caption{\color{Gray} \textbf{Visual representation of DenseNet121}, where RGB, GroundTruth Mask (viridis) \& Prediction (plasma).}
\label{fig10}
\end{figure}

\subsection{EfficientNetB0}
EfficientNetB0 is a lightweight CNN architecture designed for efficient feature extraction and image classification. It utilizes compound scaling, balancing network depth, width, and resolution to optimize performance.
\begin{figure}[H]
\centering
\includegraphics[width=0.7\textwidth]{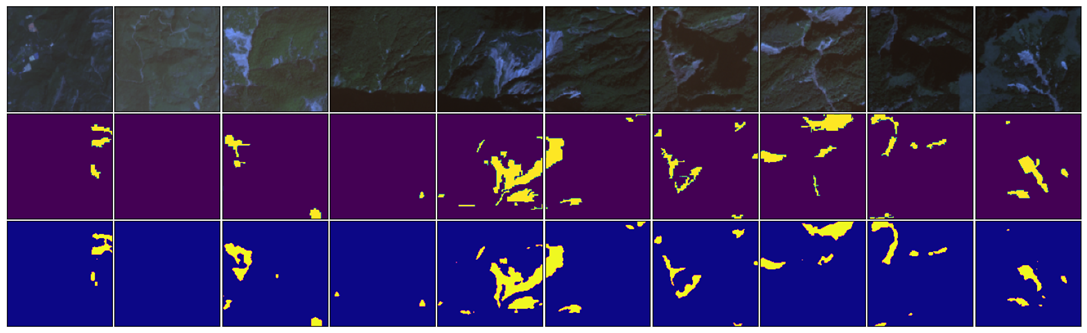}
\caption{\color{Gray} \textbf{Visual representation of EfficientNetB0}, where RGB, GroundTruth Mask (viridis) \& Prediction (plasma).}
\label{fig11}
\end{figure}

\subsection{InceptionResNetV2}
InceptionResNetV2 is a powerful deep CNN that combines the strengths of the Inception architecture and residual connections, integrating them to enhance training stability and convergence speed.
\begin{figure}[H]
\centering
\includegraphics[width=0.7\textwidth]{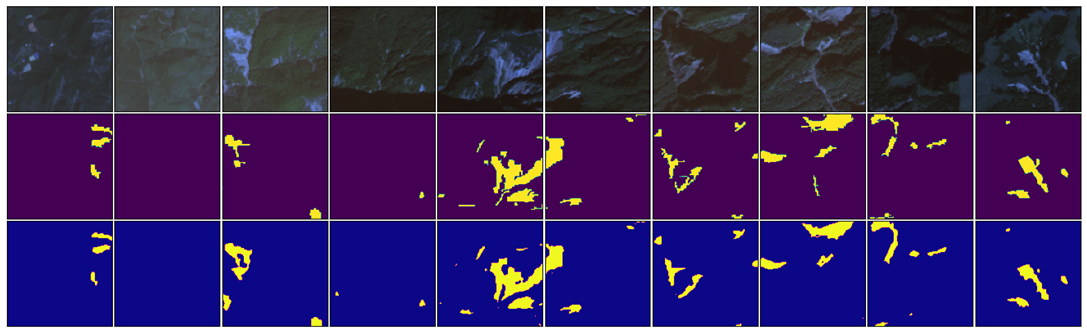}
\caption{\color{Gray} \textbf{Visual representation of InceptionResNetV2}, where RGB, GroundTruth Mask (viridis) \& Prediction (plasma).}
\label{fig12}
\end{figure}

\subsection{InceptionV4}
InceptionV4 is an advanced deep CNN that builds upon the Inception architecture by incorporating additional Inception modules and improvements in network optimization.
\begin{figure}[H]
\centering
\includegraphics[width=0.7\textwidth]{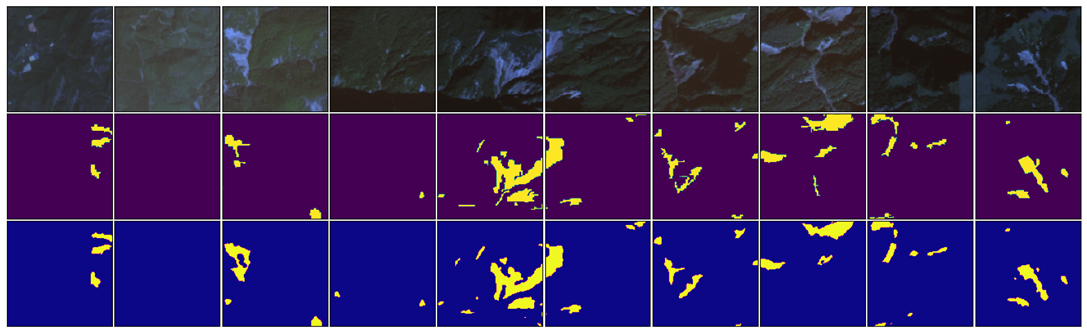}
\caption{\color{Gray} \textbf{Visual representation of InceptionV4}, where RGB, GroundTruth Mask (viridis) \& Prediction (plasma).}
\label{fig13}
\end{figure}

\subsection{MiT-B1}
MiT-B1 (Mix Transformer B1) is a lightweight yet powerful backbone used in semantic segmentation models. It employs a transformer-based approach with hierarchical feature extraction.
\begin{figure}[H]
\centering
\includegraphics[width=0.7\textwidth]{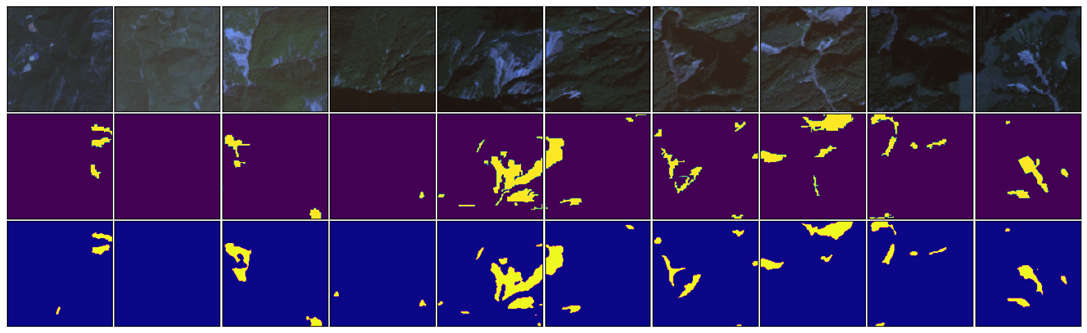}
\caption{\color{Gray} \textbf{Visual representation of MiT-B1}, where RGB, GroundTruth Mask (viridis) \& Prediction (plasma).}
\label{fig14}
\end{figure}

\newpage
\subsection{MobileNetV2}
MobileNetV2 is a lightweight and efficient deep learning model designed for mobile and edge devices, introducing an inverted residual structure and linear bottlenecks.
\begin{figure}[H]
\centering
\includegraphics[width=0.7\textwidth]{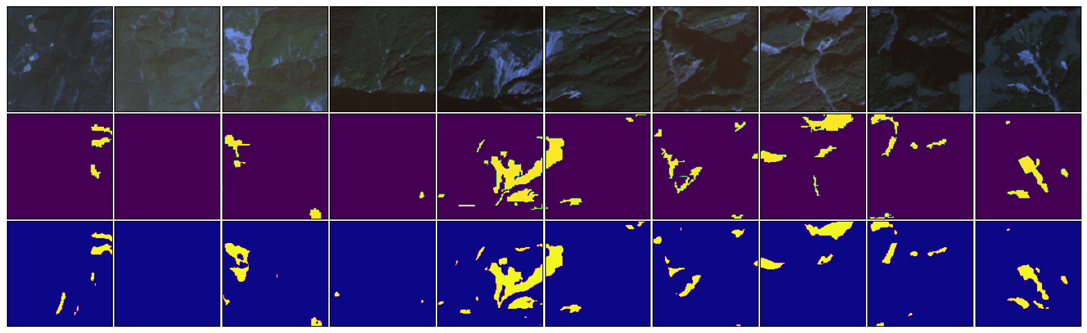}
\caption{\color{Gray} \textbf{Visual representation of MobileNetV2}, where RGB, GroundTruth Mask (viridis) \& Prediction (plasma).}
\label{fig15}
\end{figure}

\subsection{ResNet34}
ResNet34 is a deep CNN known for its residual learning framework, which helps train very deep models effectively by addressing the vanishing gradient problem.
\begin{figure}[H]
\centering
\includegraphics[width=0.7\textwidth]{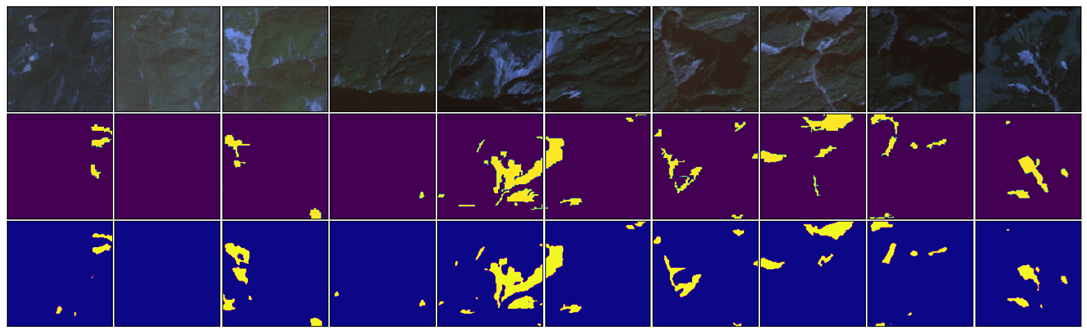}
\caption{\color{Gray} \textbf{Visual representation of ResNet34}, where RGB, GroundTruth Mask (viridis) \& Prediction (plasma).}
\label{fig16}
\end{figure}

\subsection{ResNeXt50\_32x4D}
ResNeXt50\_32x4D builds upon the ResNet architecture by incorporating a grouped convolution strategy for improved efficiency and performance.
\begin{figure}[H]
\centering
\includegraphics[width=0.7\textwidth]{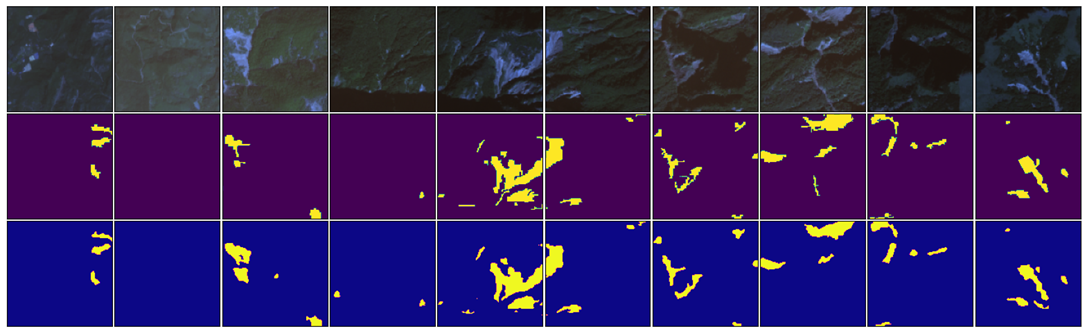}
\caption{\color{Gray} \textbf{Visual representation of ResNeXt50\_32x4D}, where RGB, GroundTruth Mask (viridis) \& Prediction (plasma).}
\label{fig17}
\end{figure}

\newpage
\subsection{SE-ResNet50}
SE-ResNet50 is an enhanced version of ResNet50 that integrates Squeeze-and-Excitation (SE) blocks to improve feature representation by modeling interdependencies between channels.
\begin{figure}[H]
\centering
\includegraphics[width=0.7\textwidth]{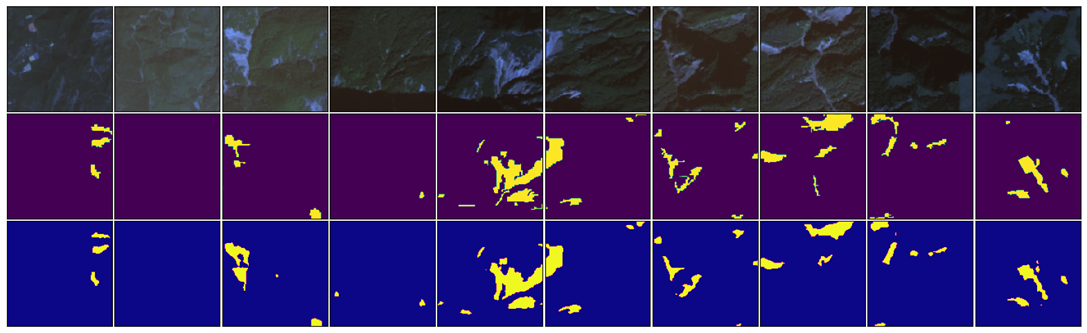}
\caption{\color{Gray} \textbf{Visual representation of SE-ResNet50}, where RGB, GroundTruth Mask (viridis) \& Prediction (plasma).}
\label{fig18}
\end{figure}

\subsection{SE-ResNeXt50-32x4D}
SE-ResNeXt50-32x4D combines ResNeXt50 with SE blocks to dynamically recalibrate channel-wise feature responses, allowing the network to focus on more relevant features.
\begin{figure}[H]
\centering
\includegraphics[width=0.7\textwidth]{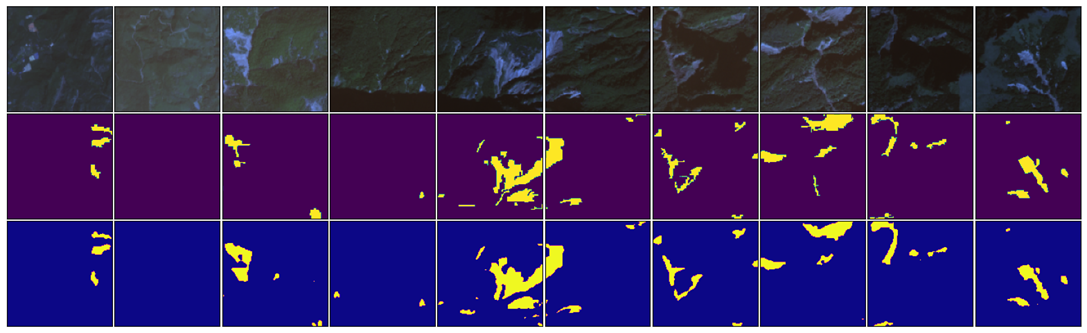}
\caption{\color{Gray} \textbf{Visual representation of SE-ResNeXt50-32x4D}, where RGB, GroundTruth Mask (viridis) \& Prediction (plasma).}
\label{fig19}
\end{figure}

\subsection{VGG16}
VGG16 is a deep CNN architecture known for its simplicity and effectiveness, consisting of 16 layers made up of small $3\times3$ convolutional filters.
\begin{figure}[H]
\centering
\includegraphics[width=0.7\textwidth]{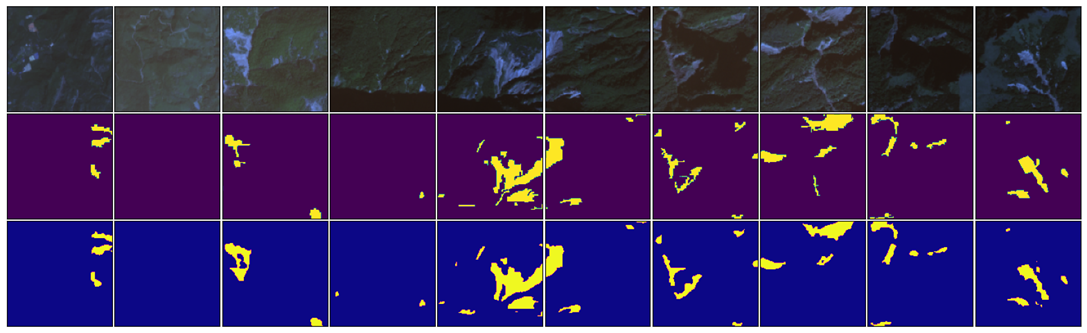}
\caption{\color{Gray} \textbf{Visual representation of VGG16}, where RGB, GroundTruth Mask (viridis) \& Prediction (plasma).}
\label{fig20}
\end{figure}

\newpage
\subsection{Loss Functions and Evaluation Metrics}
To optimize the segmentation models and evaluate their performance, we utilize a combination of loss functions and standard evaluation metrics consistently across all models. The models are trained using a combined loss function that integrates Weighted Binary Cross-Entropy (WCE) and Dice Loss to effectively handle class imbalance in landslide datasets.

\textbf{Loss Functions:}
\begin{itemize}
    \item \textbf{Weighted Binary Cross-Entropy (WCE) Loss}:
    \begin{equation}
    L_{WCE} = -w [y \log(\hat{y}) + (1-y)\log(1-\hat{y})]
    \end{equation}
    where $y$ is the ground truth, $\hat{y}$ is the predicted probability, and $w$ is the positive class weight.
    \item \textbf{Dice Loss}:
    \begin{equation}
    L_{Dice} = 1 - \frac{2 \sum (y \hat{y}) + \epsilon}{\sum y + \sum \hat{y} + \epsilon}
    \end{equation}
    where $\epsilon = 10^{-6}$ is a smoothing factor to prevent division by zero.
    \item \textbf{Combined Loss}:
    \begin{equation}
    L_{total} = (1 - \alpha) L_{WCE} + \alpha L_{Dice}
    \end{equation}
    where $\alpha = 0.5$ balances the two loss components.
\end{itemize}

\textbf{Evaluation Metrics:}
We evaluate the models using Precision, Recall, F1-Score, and Intersection over Union (IoU):
\begin{equation}
\text{Precision} = \frac{TP}{TP + FP}, \quad \text{Recall} = \frac{TP}{TP + FN}
\end{equation}
\begin{equation}
F1 = \frac{2 \cdot \text{Precision} \cdot \text{Recall}}{\text{Precision} + \text{Recall}}, \quad \text{IoU} = \frac{TP}{TP + FP + FN}
\end{equation}
where $TP$, $FP$, and $FN$ represent true positives, false positives, and false negatives, respectively.

\section{Quantitative Evaluation}
Landslide detection performance was evaluated using key segmentation metrics: F1 Score, Precision, and Recall. These metrics were derived by analyzing the pixel-wise classification results into True Positives (TP), False Positives (FP), False Negatives (FN), and True Negatives (TN). Given the critical nature of landslide prediction, an emphasis was placed on F1 Score, which balances precision and recall.

\begin{figure}[H]
\centering
\includegraphics[width=0.5\textwidth]{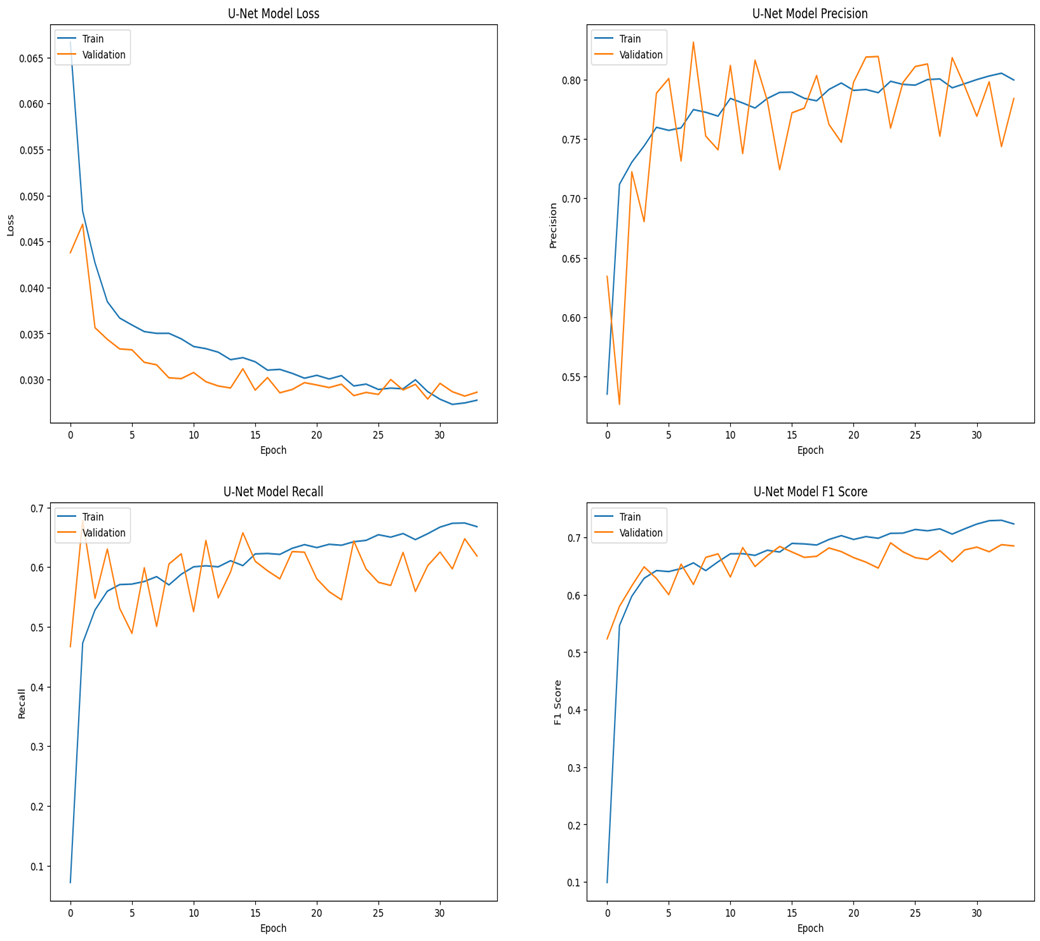}
\caption{\color{Gray} \textbf{Performance Metrics for Landslide Prediction in U-Net.}}
\label{fig21}
\end{figure}

\begin{figure}[H]
\centering
\includegraphics[width=0.48\textwidth]{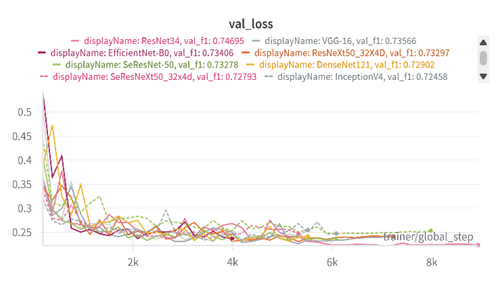} \hfill
\includegraphics[width=0.48\textwidth]{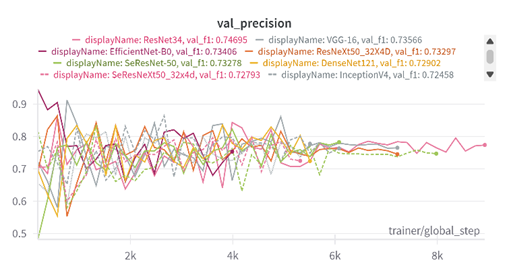} \\
\vspace{10pt}
\includegraphics[width=0.48\textwidth]{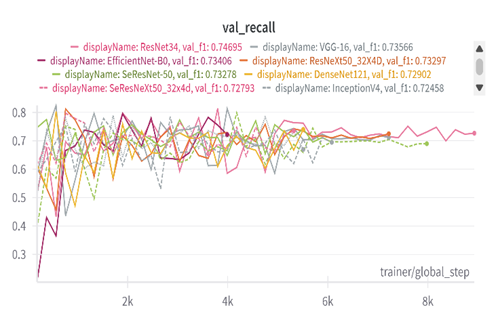} \hfill
\includegraphics[width=0.48\textwidth]{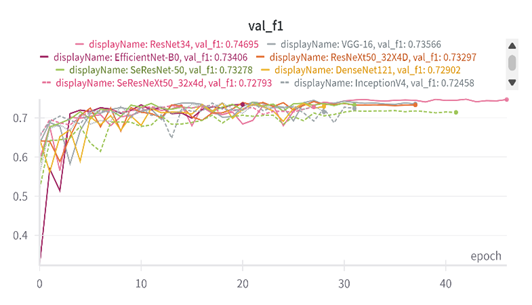}
\caption{\color{Gray} \textbf{Performance Metrics for Landslide Prediction} in ResNet34, VGG-16, EfficientNet-B0, ResNeXt50\_32X4D, SeResNet-50, DenseNet121, SeResNeXt50\_32x4D, InceptionV4, InceptionResNetV2, DeepLabV3+, MobileNetV2, MiT-b1\_14C.}
\label{fig22}
\end{figure}

\begin{table}[H]
\centering
\caption{{\bf Comparison of performance evaluation metrics of segmentation models tested on the Landslide4Sense dataset.}}
\begin{tabular}{|l|c|c|c|}
\hline
\textbf{Models} & \textbf{F1 Score} & \textbf{Precision} & \textbf{Recall} \\ \hline
ResNet34 & 0.7470 & 0.7737 & 0.7267 \\ \hline
VGG16 & 0.7357 & 0.7650 & 0.7121 \\ \hline
EfficientNet-B0 & 0.7341 & 0.7536 & 0.7221 \\ \hline
ResNeXt50\_32X4D & 0.7330 & 0.7453 & 0.7247 \\ \hline
SeResNet-50 & 0.7328 & 0.7826 & 0.6950 \\ \hline
DenseNet121 & 0.7290 & 0.7241 & 0.7400 \\ \hline
SeResNeXt50\_32x4D & 0.7279 & 0.7249 & 0.7350 \\ \hline
InceptionV4 & 0.7246 & 0.7631 & 0.6945 \\ \hline
InceptionResNetV2 & 0.7151 & 0.7774 & 0.6692 \\ \hline
DeepLabV3+ & 0.7141 & 0.7471 & 0.6897 \\ \hline
MobileNetV2 & 0.7119 & 0.7000 & 0.7337 \\ \hline
U-Net & 0.7012 & 0.7906 & 0.6338 \\ \hline
MiT-B1 & 0.6989 & 0.7574 & 0.6596 \\ \hline
\end{tabular}
\label{tab2}
\end{table}

Our results demonstrate that ResNet34, VGG-16, and EfficientNet-B0 achieved the highest F1 Scores, indicating superior performance in distinguishing landslide-prone areas from non-landslide regions. The ResNet34-based U-Net model attained the best balance between precision and recall, achieving an F1 Score of 0.7470, making it the most reliable among the tested architectures.

The classic U-Net architecture, while still effective, demonstrated a lower F1 Score of 0.7012, highlighting the advantage of deeper and more advanced feature extraction architectures like Res-Net and EfficientNet-B0.

\section{Results}
In this study, we systematically evaluated the performance of various deep learning models for landslide detection and segmentation, utilizing the Landslide4Sense dataset. Among all models tested, the U-Net architecture with a ResNet34 backbone achieved the highest F1 Score of 0.7470, indicating its superior performance in distinguishing landslide-prone areas from non-landslide regions. This model exhibited a balanced trade-off between precision (0.7737) and recall (0.7267), suggesting that it effectively captures both true positives and minimizes false negatives. 

Other models such as ResNeXt50\_32X4D, SeResNet-50, and DenseNet121 showcased competitive performance, further highlighting the benefits of integrating deeper and more complex feature extraction mechanisms. These models, although not reaching the F1 Score of ResNet34-based U-Net, still performed well, particularly in identifying nuanced features in the satellite imagery. Overall, these results underscore the importance of using advanced deep learning models and multi-source satellite data to enhance landslide detection performance.

\section{Conclusion}
In conclusion, the findings from this study provide strong evidence that hybrid deep learning models leveraging deeper feature extraction mechanisms significantly improve landslide detection performance compared to traditional U-Net. The ResNet34-based U-Net architecture emerged as the most reliable model, achieving the highest F1 Score of 0.7470, which reflects its ability to maintain a strong balance between precision and recall. 

The results of this study provide valuable insights into the development of more reliable and scalable landslide detection systems, which are crucial for disaster risk management, early warning systems, and environmental monitoring in landslide-prone regions. The ability to incorporate both local and global contextual information from various data sources is critical for improving segmentation quality and prediction accuracy. 

\nolinenumbers

\addcontentsline{toc}{section}{References}
\bibliography{library}
\bibliographystyle{abbrv}

\end{document}